\documentclass[10pt, a4paper]{article}
\usepackage[utf8]{inputenc}
\usepackage{authblk}
\usepackage[final]{lrec2026}
\usepackage{amsmath}
\usepackage{amsfonts} 
\usepackage{tipa}
\usepackage{graphicx}
\usepackage{hyperref}
\usepackage{newunicodechar}
\newunicodechar{ʿ}{'}

\title{CVPD at QIAS 2026: RAG-Guided LLM Reasoning for Al-Mawarith Share Calculation and Heir Allocation}

\name{{\small\bfseries Wassim Swaileh\textsuperscript{1,4,*}, Mohammed-En-Nadhir Zighem\textsuperscript{2,*},}\\
{\small\bfseries Hichem Telli\textsuperscript{2,*}, Salah Eddine Bekhouche\textsuperscript{2,3,*}, Abdellah Zakaria Sellam\textsuperscript{4},}\\
{\small\bfseries Fadi Dornaika\textsuperscript{3,5} and Dimitrios Kotzinos\textsuperscript{1}}}

\address{
\textsuperscript{1} ETIS (UMR 8051), CY Cergy Paris Univ., ENSEA, CNRS, France \\
\textsuperscript{2} VSC Laboratory, Department of Electronics and Automation,
University of Biskra, Algeria \\
\textsuperscript{3} Dept. of CS \& AI, Univ. of the Basque Country (UPV/EHU), Spain \\
\textsuperscript{4} Computer Engineering Dept, Sana'a Community College, Yemen \\
\textsuperscript{5} CNR-ISASI "E. Caianiello", Lecce, Italy \\
\textsuperscript{6} Ikerbasque, Bilbao, Spain \\
{\footnotesize\texttt{\{wassim.swaileh,dimitris.kotzinos\}@ensea.fr}}\\
{\footnotesize\texttt{\{mohammedennadhir.zighem,tellihicham\}@univ-biskra.dz}}\\
{\footnotesize\texttt{\{salaheddine.bekhouche,fadi.dornaika\}@ehu.eus}}\\
{\footnotesize\texttt{abdellahzakaria.sellam@cnr.it}}
}

\abstract{
Islamic inheritance law involves multi-stage legal reasoning, including identifying eligible heirs, resolving blocking, assigning fixed and residual shares, handling ‘awl and radd adjustments, and producing a consistent post-taʿsīl distribution. Variations across madhāhib and national civil codes complicate this task, requiring models to rely on explicit legal rules rather than implicit knowledge. We introduce a retrieval-augmented generation pipeline that combines rule-grounded synthetic data, hybrid retrieval with cross-encoder reranking, and schema-constrained output validation. Our system achieved a 0.935 MIR-E score and ranked first on the QIAS 2026 blind-test leaderboard, demonstrating that retrieval-grounded, schema-aware generation excels in high-precision Arabic legal reasoning.
\newline  }

\begin{document}

\maketitleabstract

\section{Introduction}

Islamic inheritance law demands precise multi-stage reasoning: identifying eligible heirs and blockings, assigning fixed and residual shares, applying ‘awl and radd adjustments, and computing post-ta'sīl distributions \cite{ahmad2025transformer}. Rules vary across madhāhib and civil codes, requiring explicit configurations rather than general knowledge. QIAS 2026 \cite{bouchekif2025qias} evaluates systems via MIR-E, scoring intermediate stages and final outputs where structure matches legal correctness in importance. Large language models struggle with such multi-step reasoning and rule adherence in domains like Islamic jurisprudence \cite{bekhouche2025ars}\cite{bouchekif2025qias}.
We cast the task as \emph{retrieval-augmented structured prediction} and propose a three-component pipeline:

(i)~a \textbf{deterministic rule-based generator and symbolic inheritance calculator} that constructs a large synthetic legal corpus with full stage-wise reasoning traces;

(ii)~a \textbf{hybrid retriever (semantic + BM25) with cross-encoder reranking} for selecting the most relevant solved cases as generation context; and

(iii)~a \textbf{constrained LLM decoding-and-parsing stage} with a downstream validation layer enforcing type, key, and cross-field invariants.


This pipeline achieved a MIR-E score of 0.935 and ranked first on the QIAS 2026 blind-test leaderboard\footnote{https://sites.google.com/view/qias2026/}, confirming that retrieval-grounded, schema-aware generation is effective for high-precision Arabic legal reasoning. The full pipeline is available for public use at \href{https://github.com/swaileh/qias-mawarith-rag}{qias-mawarith-rag repository}

\section{Related Work}

The QIAS shared tasks \cite{bouchekif2025qias} established Islamic inheritance reasoning as a benchmark for Arabic legal AI, requiring both legal interpretation and precise multi-stage numerical computation. Prior work explored retrieval-augmented generation \cite{ahmad2025transformer}, multi-agent and fine-tuning approaches \cite{phuc2025puxai,mohammad2025qu}, Arabic retrieval optimization \cite{bekhouche2025ars}, encoder-based inheritance reasoning \cite{bekhouche2025cvpd}, and broader LLM benchmarking \cite{aldahoul2025nyuad}. While retrieval and ensemble methods show promise \cite{hamed2025hiast,r2025morai}, common augmentation techniques do not ensure legal or arithmetic correctness. Recent QIAS studies further show that even strong LLMs struggle with multi-step mathematical reasoning, strict stage-wise formatting, and numerical consistency \cite{aldahoul2025nyuad,ahmad2025transformer}. Our work differs in two ways: we improve the data distribution through rule-grounded synthetic generation with full intermediate traces, and emphasize deterministic validation around generation for end-to-end MIR-E evaluability.

\section{Method}
\begin{figure}[htbp] 
    \centering
    \includegraphics[width=0.45\textwidth]{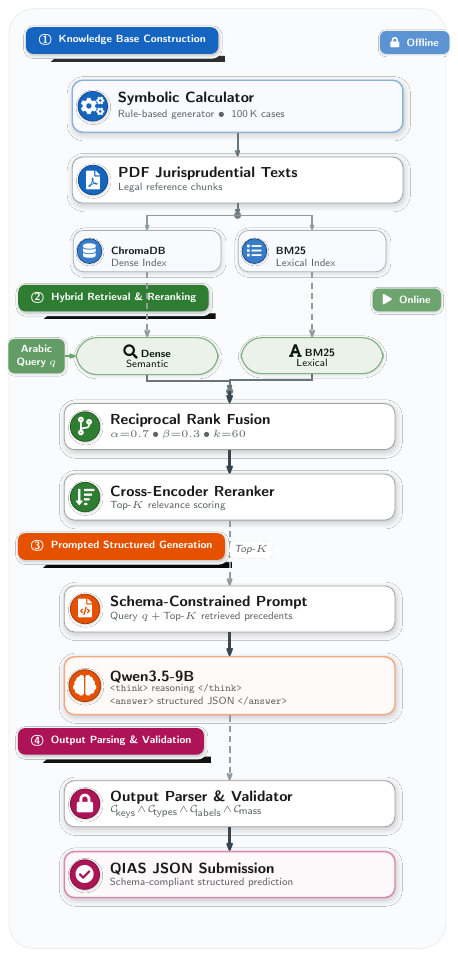}
    \caption{Overview of the proposed RAG pipeline for QIAS.}
    \label{fig:Fig1}
\end{figure}

\subsection{Task Setting and Data Profile}

The QIAS 2026 dataset comprises 5,900 training examples, 200 development examples   . Each includes an Arabic scenario alongside a structured gold output listing heirs, blocked heirs, shares, adjustment status, and final post-ta'sīl distribution. The dataset reveals a train-dev mismatch: the training set favors simple cases, while the development set features more Radd scenarios. This imbalance motivates retrieval from a broader synthetic corpus rather than relying solely on parametric memorization. Table~\ref{tab:dataset} summarizes the label distributions.
\begin{table}[h]
\centering
\small
\begin{tabular}{@{}lccc@{}}
\hline
\textbf{Category} & \textbf{Train} & \textbf{Dev} & \textbf{Diff} \\
\hline
Simple & 5,531 (93.8\%) & 125 (62.5\%) & -31.3\% \\
`Awl & 288 (4.9\%) & 4 (2.0\%) & -2.9\% \\
Radd & 81 (1.4\%) & 71 (35.5\%) & +34.1\% \\
\hline
\textbf{Total} & 5,900 & 200 & -- \\
\hline
\end{tabular}
\caption{QIAS 2026 dataset distribution showing train--dev imbalance.}
\label{tab:dataset}
\end{table}

\subsection{Rule-Based Generator and Symbolic Calculator}

To obtain high-fidelity supervision, we adopt a \emph{symbolic-first} data synthesis strategy. For each sampled heir configuration, a deterministic inheritance calculator resolves all QIAS stages in sequence, eligibility and blocking, fixed and residual share assignment, global adjustment, and ta\c{s}={\i}l---prior to any natural-language generation. This ensures jurisprudential validity and arithmetic consistency by construction. Scenario generation covers spouses, descendants, ascendants, siblings, and extended family under controllable difficulty regimes. To boost coverage of rare but evaluation-critical cases, we inject targeted templates (e.g., `Awl-heavy and Radd-heavy configurations) and enforce uniqueness over heir combinations to minimize duplication. The pipeline yields $\sim$100,000 unique synthetic cases, each rendered in four complementary views: Arabic problem text, natural-language Q\&A form, explicit reasoning trace, and full structured output. The Q\&A rendering serves as the primary retrievable document for indexing.

\subsection{RAG Pipeline}

\subsubsection{Hybrid Retrieval and Reranking}

At inference time, retrieval combines dense semantic search and sparse BM25 lexical search.
Let $q$ be the query and $d_i$ a candidate document.
Dense retrieval uses embedding similarity $s_{\mathrm{dense}}(q,d_i)=\cos(E(q),E(d_i))$.
Candidate sets from both channels are fused using weighted Reciprocal Rank Fusion:
\begin{equation}
\mathrm{RRF}(d_i)
=\alpha\frac{1}{k+r_{\mathrm{dense}}(d_i)}
+\beta\frac{1}{k+r_{\mathrm{bm25}}(d_i)}
\end{equation}
with $(\alpha,\beta)=(0.7,0.3)$ and smoothing constant $k=60$.
The top fused set is reranked by a cross-encoder score $s_{\mathrm{ce}}(q,d_i)$ and truncated to the final context set $\mathcal{D}_K=\operatorname{TopK}_{d_i}\, s_{\mathrm{ce}}(q,d_i)$.

\subsubsection{Prompted Structured Generation}

Retrieved evidence is injected into an Arabic prompt requesting decomposition into reasoning (\texttt{<think>}) and final answer (\texttt{<answer>}).
Given context $\mathcal{D}_K$, generation follows:
\begin{equation}
y^*=\arg\max_y P_\theta(y \mid q,\mathcal{D}_K),
\end{equation}
with an operational decomposition into reasoning and structured output:
\begin{equation}
\begin{aligned}
P_\theta(y\mid q,\mathcal{D}_K)
&=P_\theta(y_{\texttt{think}}\mid q,\mathcal{D}_K)\cdot{}\\
&\quad P_\theta(y_{\texttt{answer}}\mid q,\mathcal{D}_K,y_{\texttt{think}}).
\end{aligned}
\end{equation}
We used Qwen3.5 - 9B LLM model \cite{qwen35reprt2026} to produced the structured output. The model is encouraged to produce explicit intermediate justification while preserving a machine-readable final structure.

\subsubsection{Output Parsing and Validation}
A multi-stage post-processing module conducts structured information extraction, eliminates generation artifacts, and deploys controlled fallback strategies in cases where explicit JSON output is absent. The validation procedure evaluates four categories of constraints:
\begin{equation}
\mathcal{V}(y)=
\mathcal{C}_{\mathrm{keys}}
\wedge
\mathcal{C}_{\mathrm{types}}
\wedge
\mathcal{C}_{\mathrm{labels}}
\wedge
\mathcal{C}_{\mathrm{mass}}.
\end{equation}
Specifically, $\mathcal{C}_{\mathrm{keys}}$ verifies the presence of critical schema keys (\texttt{heirs}, \texttt{shares}, \texttt{awl\_or\_radd});
$\mathcal{C}_{\mathrm{types}}$ ensures that all fields conform to their prescribed data types;
$\mathcal{C}_{\mathrm{labels}}$ validates that categorical fields assume permissible legal values;
and $\mathcal{C}_{\mathrm{mass}}$ enforces coherence of the post-ta\c{s}\={\i}l share distribution.
For non-critical missing keys, rule-based default values are introduced to enhance the completeness and usability of the submissions.

\section{Evaluations and Results}
\subsection{Evaluation Protocol}
Evaluation follows the official QIAS 2026 protocol \cite{bouchekif2025qias}:
predictions are scored at each intermediate reasoning stage (heirs, blocked, shares, `Awl/Radd, final distribution) and aggregated into a weighted MIR-E score.
Set-valued stages use order-invariant F1, while value stages use indicator scoring.
Full scoring definitions and stage weights are given in Appendix~\ref{app:mire}.
All results are evaluated on the 200-case development split and 500-case test split; final ranking uses the official blind-test leaderboard.

\subsection{Off-the-Shelf Benchmark}
Before full pipeline optimization, we benchmarked off-the-shelf models to select a strong open-source base.
Results in Table~\ref{tab:benchmark} show that Qwen 3.5 9B provides the strongest initial reasoning among open models, while output completeness remains a bottleneck.
This observation is consistent with prior QIAS studies reporting useful legal reasoning but recurring failures in structured-output reliability \cite{aldahoul2025nyuad,ahmad2025transformer}.
This trade-off motivates our design choice: start from the strongest open reasoning prior, then improve structural reliability through retrieval grounding and post-generation validation.
\begin{table}[t]
\centering
\small
\begin{tabular}{@{}lcc@{}}
\hline
\textbf{Model} & \textbf{MIR-E} & \textbf{Missing} \\
\hline
Qwen 3.5 9B & 0.347 & 19  \\
Qwen2.5-7B     & 0.243 & 0   \\
Command-R7B    & 0.225 & 32  \\
Llama3.2-1B    & 0.109 & 80  \\
Gemma3-12B     & 0.010 & 193 \\
\hline
\end{tabular}
\caption{Off-the-shelf benchmarking on the 200-case QIAS dev set.}
\label{tab:benchmark}
\end{table}
\subsection{Official Blind-Test Ranking}

On the official QIAS 2026 blind test, the system ranked first overall.
Table~\ref{tab:official_ranking} summarizes leaderboard performance.
The narrow margin at the top (0.935 vs.\ 0.931) indicates a highly competitive setting.
Our rank-1 result supports the practical robustness of combining curated legal retrieval with deterministic post-generation validation.
Relative to prior QIAS designs based on retrieval prompting, multi-agent orchestration, or direct fine-tuning \cite{ahmad2025transformer,phuc2025puxai,mohammad2025qu}, the main finding is that curated legal retrieval with deterministic validation improves both competitiveness and submission stability under MIR-E.
\begin{table}[t]
\centering
\small
\begin{tabular}{@{}ccc@{}}
\hline
\textbf{Rank} & \textbf{Team} & \textbf{MIR-E} \\
\hline
1 & \textbf{CVPD (ours)} & 0.935 \\
2 & Simplicity \cite{almansour2026simplicity} & 0.931 \\
3 & KMS \cite{alkhamis2026kms} & 0.916 \\
4 & QU-NLP\textsuperscript{*} & 0.907 \\
5 & PSL \cite{mouhoub2026psl} & 0.898 \\
6 & grkurdi\textsuperscript{*} & 0.826 \\
7 & UTLM\textsuperscript{*} & 0.742 \\
8 & rouba1234\textsuperscript{*} & 0.325 \\
\hline
\end{tabular}
\caption{Official QIAS 2026 blind-test leaderboard. Teams marked with * did not have published system descriptions at the time of writing.}
\label{tab:official_ranking}
\end{table}

\subsection{Ablation}
We perform a controlled source ablation to isolate how evidence origin affects retrieval quality.
The study is run on the full QIAS development set (200 questions) under three retrieval settings:
\textbf{PDF-only}, \textbf{Web-only}, and \textbf{PDF+Web}.
We report retrieval-side metrics:
mean semantic similarity, mean keyword overlap, and a weighted combined score
$S_{\mathrm{comb}} = 0.5\,S_{\mathrm{sem}} + 0.3\,S_{\mathrm{kw}} + 0.2\,S_{\mathrm{tfidf}}$.
Retrieval success is defined as $S_{\mathrm{comb}} \geq 0.5$.
PDF-only achieves the highest combined score (0.718), outperforming PDF+Web (0.591) and Web-only (0.463), with a best--worst gap of 0.254.
The same ordering holds for semantic similarity (0.879, 0.746, 0.593), indicating that legal coherence of retrieved context is primarily driven by curated textual evidence.
Web-only shows slightly higher keyword overlap (0.269) but this does not translate into robust retrieval quality, with retrieval success dropping to 22.5\% vs.\ 97.5--100\% for PDF-containing settings.
PDF-only produces 190/200 high-quality retrievals and no poor cases, while Web-only concentrates on lower-quality bins (176 fair, 23 poor).
These findings support using curated PDF knowledge as the primary retrieval backbone, consistent with broader legal-NLP evidence that domain-bounded corpora provide higher semantic precision than open web text for high-stakes reasoning tasks.
\begin{table}[h]
\centering
\small
\begin{tabular}{@{}lccc@{}}
\hline
\textbf{Source} & \textbf{Combined} & \textbf{Semantic} & \textbf{Keyword} \\
\hline
PDF-only & 0.718 & 0.879 & 0.235 \\
PDF+Web & 0.591 & 0.746 & 0.229 \\
Web-only & 0.463 & 0.593 & 0.269 \\
\hline
\end{tabular}
\caption{Source ablation on the 200-question QIAS dev set.}
\label{tab:ablation_sources}
\end{table}

\subsection{Error Analysis}

Despite strong aggregate performance, residual failures are concentrated in a few high-impact categories:
\begin{itemize}
    \item \textbf{Blocking logic (42\%)}: multi-generation \textit{hajb} interactions remain the dominant error source.
    \item \textbf{Fraction arithmetic (31\%)}: uncommon denominator interactions still trigger occasional numeric drift.
    \item \textbf{`Awl handling (18\%)}: edge cases with tightly coupled reductions are underrepresented and difficult.
    \item \textbf{Formatting failures (9\%)}: long generations can still induce partial schema violations.
\end{itemize}

These patterns are consistent with the task structure: errors arise less from surface language understanding and more from compositional legal-arithmetic dependencies.
Future improvements should therefore prioritize explicit blocking-graph reasoning and targeted training on rare denominator/`Awl templates.

\section{Discussion}

Our results yield three main findings.
First, strong base LLMs remain inadequate without structural support: modern instruction-tuned models generate plausible legal reasoning but still fail strict stage-wise consistency \cite{aldahoul2025nyuad,ahmad2025transformer}.
Second, retrieval-grounded, schema-constrained generation is a robust improvement path; compared with earlier QIAS pipelines \cite{ahmad2025transformer,phuc2025puxai,mohammad2025qu,bekhouche2025cvpd}, reliability-oriented generation engineering is as important as model choice.
Third, curated PDF jurisprudential evidence consistently beats Web-only retrieval in semantic quality, confirming that domain-bounded corpora better support high-precision legal reasoning \cite{hamed2025hiast,r2025morai}.
Error analysis shows that remaining bottlenecks are compositional legal–arithmetic dependencies, not surface-level Arabic understanding.
Despite competitiveness, several limitations remain:

\begin{itemize}
    \item \textbf{Jurisprudential coverage}: current rule profiles do not cover all major Sunni schools and civil-law variants.
    \item \textbf{Source sensitivity}: Web evidence can harm semantic precision when fusion is loosely controlled.
    \item \textbf{Residual reasoning errors}: multi-generation \textit{hajb} dependencies and rare denominator interactions remain challenging.
    \item \textbf{Evaluation scope}: MIR-E only partially captures explanation faithfulness and practical usability.
\end{itemize}

\noindent Future work will broaden symbolic profiles to more \textit{madh\=ahib}, add confidence-aware retrieval with source attribution, integrate blocking-graph constraints for neuro-symbolic error reduction, and augment MIR-E with expert juristic review.

\section{Conclusion}

We presented a retrieval-augmented, schema-constrained pipeline for Islamic inheritance reasoning at QIAS 2026, combining symbolic data synthesis (100,000 cases), hybrid retrieval with cross-encoder reranking, and deterministic output validation.
The system achieved 0.935 MIR-E and ranked first on the official blind-test leaderboard, showing that progress on Arabic legal reasoning depends on explicit control of evidence quality, stage-level consistency, and schema compliance.

\section{Bibliographical References}
\label{sec:reference}

\bibliographystyle{lrec2026-natbib}
\bibliography{references}


\newpage
\appendix

\begin{center}
\textbf{\Large Appendix: Supplementary Material}
\end{center}

\vspace{0.5em}

This appendix gives additional technical details and analysis that could not be included in the main paper due to space limits. We cover the MIR-E scoring protocol (Section~\ref{app:mire}), the full BM25 formulation (Section~\ref{app:bm25}), more details about output validation (Section~\ref{app:validation}), the submission format (Section~\ref{app:submission}), a deeper ablation analysis (Section~\ref{app:ablation_extended}), and an extended discussion (Section~\ref{app:discussion}).

\section{MIR-E Scoring Protocol}
\label{app:mire}

The evaluated stages are:
\texttt{heirs}, \texttt{blocked}, \texttt{shares}, \texttt{awl\_or\_radd}, \texttt{awl\_stage}, and \texttt{post\_tasil}.

For set-valued stages (eligible heirs and blocked heirs), scoring is order-invariant:
\begin{equation}
S_{\mathrm{set}}=\frac{2\cdot |X_{\mathrm{gold}}\cap X_{\mathrm{pred}}|}{|X_{\mathrm{gold}}|+|X_{\mathrm{pred}}|}.
\end{equation}
For value-based stages, scoring is indicator-based with tolerance $\varepsilon$:
\begin{equation}
S_{\mathrm{value}}=
\begin{cases}
1, & |v_{\mathrm{pred}}-v_{\mathrm{gold}}|\le \varepsilon \\
0, & \text{otherwise.}
\end{cases}
\end{equation}

Stage weights are:
$\lambda_1{=}0.30$ (heirs),
$\lambda_2{=}0.20$ (blocked),
$\lambda_3{=}0.20$ (shares),
$\lambda_4{=}0.10$ (`Awl/\textit{Radd}),
$\lambda_5{=}0.20$ (final distribution).
The overall MIR-E score is:
\begin{multline}
\mathrm{MIR\text{-}E}=
\lambda_1 S_{\mathrm{heirs}}+
\lambda_2 S_{\mathrm{blocked}}+
\lambda_3 S_{\mathrm{shares}}\\
+\,\lambda_4 S_{\mathrm{awl}}+
\lambda_5 S_{\mathrm{final}}.
\end{multline}

\section{Full BM25 Scoring Formula}
\label{app:bm25}

In the main paper, we mention that the sparse retrieval channel relies on BM25 but we do not show the formula in detail. The full expression is:
\begin{multline}
s_{\mathrm{bm25}}(q,d_i)
=\sum_{t\in q}\mathrm{IDF}(t)\cdot{}\\
\frac{f(t,d_i)(k_1+1)}
{f(t,d_i)+k_1\bigl(1-b+b\,\tfrac{|d_i|}{\mathrm{avgdl}}\bigr)}.
\end{multline}
$f(t,d_i)$ stands for the term frequency of $t$ in document $d_i$, $|d_i|$ is how long the document is, and $\mathrm{avgdl}$ is the mean document length over the whole corpus. We set $k_1=1.2$ and $b=0.75$, which are standard choices.
This score gives more weight to terms that are rare across documents (through IDF) while normalizing for document length.

\section{More Details on Output Validation}
\label{app:validation}

Section 3.2.3 of the main paper describes the validation formula with four constraints. Here we give some extra implementation details.

Not all keys in the output JSON carry the same importance. When a non-critical field is missing (for example \texttt{blocked} or \texttt{tasil\_stage}), the validator fills in a default value instead of rejecting the whole prediction. This design choice helped a lot with submission completeness, especially for cases where the model produces good reasoning but drops some fields.

For share coherence, we check whether the total of all heir percentages sums to a value close to 100:
\begin{equation}
|\sum_{h\in\mathcal{H}} p_h \cdot c_h - 100|\le \varepsilon,\quad \varepsilon=5,
\end{equation}
where $p_h$ is the per-head post-ta\c{s}\={\i}l percentage for heir $h$ and $c_h$ is the count. We allow a tolerance of $\varepsilon=5$ to handle rounding effects that come from fraction-based share divisions.

We apply this same protocol (extraction, cleanup, fallback, then validation) for both the development set runs and the test set predictions, so the results are directly comparable.

\section{Submission Format}
\label{app:submission}

The official submission format requires a single file called \texttt{submission.json}, packed inside a ZIP archive. Each entry in this file has three fields: \texttt{id}, \texttt{question}, and a structured \texttt{output} object that covers all the reasoning stages.

One thing worth noting: in our pipeline code we use the variable name \texttt{tasil\_stage} for the intermediate adjustment state. The organizer's evaluator expects a slightly different name, so we do a renaming step before packaging the submission. This is easy to miss and we include it here for others building on this pipeline.

\section{Source Ablation: Additional Analysis}
\label{app:ablation_extended}

The main paper shows the headline numbers for the source ablation (Table 4). Below we break the results down further by quality levels and question difficulty.

\subsection{Results by Quality Levels}

We grouped retrievals into four bins -- \emph{excellent}, \emph{good}, \emph{fair}, and \emph{poor} -- based on the combined similarity score. The pattern was quite clear.

With PDF-only, 190 out of 200 retrievals fell into the top two bins (\emph{excellent} or \emph{good}), and there were zero \emph{poor} retrievals. PDF+Web was still decent -- no \emph{poor} cases either -- but a portion of the results shifted from \emph{good} down to \emph{fair}. It seems like adding web documents broadens the context but dilutes the semantic precision.

Web-only told a different story. Most retrievals (176) ended up in the \emph{fair} category, and 23 were \emph{poor}. Almost none reached the top bins. This confirms that noisy web sources are not reliable on their own for this kind of legal task.

\subsection{Question Difficulty}

We also looked at how difficulty affects the results. Out of the 200 development questions, 61 were simple (30.5\%), 67 moderate (33.5\%), 33 complex (16.5\%), and 39 very complex (19.5\%). The average complexity score was 5.14 out of 10, so the set is reasonably diverse.

What we found is that PDF-only retrieval stays strong even on harder questions, but Web-only gets worse fast as difficulty goes up. This tells us that source quality matters more than how hard the question is -- at least for the retrieval stage of the pipeline.

\section{Extended Discussion}
\label{app:discussion}

\subsection{Key Findings in More Detail}

We highlight three main takeaways from this work.

\textbf{Off-the-shelf models are not enough.}
Our benchmark shows that modern LLMs can generate plausible legal reasoning on inheritance problems, but they struggle with strict stage-wise consistency. Missing fields and wrong output formats are common problems. This is consistent with what other teams reported in earlier QIAS editions \cite{aldahoul2025nyuad,ahmad2025transformer}. The models ``know'' the law at some level, but they cannot reliably produce all the required output fields in the correct format.

\textbf{Retrieval grounding helps, but engineering matters too.}
The gap between the top teams on the leaderboard was small (0.935 vs.\ 0.931). At this level, the choice of model matters less than what you build around it. Our experience suggests that careful post-processing, validation, and fallback logic contributed as much to the final score as the retrieval itself. Compared to earlier QIAS approaches that used retrieval prompting \cite{ahmad2025transformer}, multi-agent systems \cite{phuc2025puxai,mohammad2025qu}, or encoder-based methods \cite{bekhouche2025cvpd}, the lesson is: spending effort on making the outputs reliable can pay off more than switching to a bigger model.

\textbf{Curated sources beat web sources.}
PDF-based retrieval was consistently better than web-based retrieval across all the metrics we checked. Web sources provide slightly more keyword overlap, but their semantic quality is lower and they bring in noise. The legal-NLP community has pointed this out before \cite{hamed2025hiast,r2025morai}, and our experiments confirm it once more for Arabic inheritance law.

One last observation: the errors that remain are mostly about compositional legal reasoning (blocking chains, fraction arithmetic, rare `Awl edge cases), not about Arabic text understanding. So the next improvements should probably come from better symbolic reasoning modules, not from better language models.

\subsection{Limitations and Directions for Future Work}

There are several things we could not address in this version of the system:

\begin{itemize}
    \item \textbf{Coverage of legal schools}: Right now the rule profiles cover the main cases, but we do not yet handle all the differences between the four Sunni \textit{madh\=ahib}. Some civil-law codifications also have their own variants. Broadening this coverage is the most important next step.
    \item \textbf{Noisy web sources}: When web evidence is included without enough filtering, it can hurt results. A confidence-aware retrieval approach, where the system estimates how trustworthy each source is, could help manage this issue.
    \item \textbf{Hard reasoning cases}: Multi-generation blocking (\textit{hajb}) and uncommon denominator combinations are still the main sources of error. We think explicit symbolic constraints -- like a blocking graph or a fraction checker -- could target these specific failure patterns.
    \item \textbf{Evaluation beyond MIR-E}: The MIR-E metric checks whether each stage is correct, but it does not measure how useful or understandable the explanations are. Expert review from Islamic law scholars, together with cross-benchmark evaluation, would give a better picture of real-world readiness.
\end{itemize}

\end{document}